\documentclass[a4paper]{article}
\usepackage{idsiatechrep}      % english acronyms
\usepackage{times}
\usepackage{epsfig}
\usepackage{graphicx}
\usepackage{amsmath}
\usepackage{amssymb}
\usepackage{subfigure}
\usepackage[pagebackref=true,breaklinks=true,letterpaper=true,colorlinks,bookmarks=false]{hyperref}

\title{Multi-column Deep Neural Networks for Image Classification}
\author{Dan Cire\c{s}an\\
  Ueli Meier\\
J{\"u}rgen Schmidhuber}
\date{February 2012}

\reportnumber{04-12}

\begin{document}
\makecover         % makes the cover sheet
\maketitle

\begin{abstract}
Traditional methods of computer vision and machine learning cannot match human performance on tasks such as the recognition of handwritten digits or traffic signs. Our biologically plausible deep artificial neural network architectures can.
Small (often minimal) receptive fields of convolutional winner-take-all neurons yield large network depth, resulting in roughly as many sparsely connected neural layers as found in mammals between retina and visual cortex. Only winner neurons are trained. Several deep neural columns become experts on inputs preprocessed in different ways; their predictions are averaged. Graphics cards allow for fast training. On the very competitive MNIST handwriting benchmark, our method is the first to achieve near-human performance. On a traffic sign recognition benchmark it outperforms humans by a factor of two. We also improve the state-of-the-art on a plethora of common image classification benchmarks.
\end{abstract}

\section{Introduction}

Recent publications suggest that unsupervised pre-training of deep, hierarchical neural networks improves supervised pattern classification  \cite{bengio:2007,erhan:2010}. Here we train such nets by simple online back-propagation, setting new, greatly improved records on MNIST \cite{lecun:1998}, Latin letters \cite{NIST}, Chinese characters \cite{cheng-lin:2010}, traffic signs \cite{stallkamp:2011}, NORB (jittered, cluttered) \cite{lecun:2004} and CIFAR10 \cite{krizhevsky:2009} benchmarks.

We focus on deep convolutional neural networks (DNN), introduced by \cite{fukushima:1980}, improved by \cite{lecun:1998}, refined and simplified by \cite{behnke:2003,simard:2003,Ciresan:2011a}. Lately, DNN proved their mettle on data sets ranging from handwritten digits (MNIST) \cite{Ciresan:2010,Ciresan:2011a}, handwritten characters \cite{Ciresan:2011c} to 3D toys (NORB) and faces \cite{strigl:2010}. DNNs fully unfold their potential when they are big and deep \cite{Ciresan:2011a}. But training them requires weeks, months, even years on CPUs. High data transfer latency prevents multi-threading and multi-CPU code from saving the situation. In recent years, however, fast parallel neural net code for graphics cards (GPUs) has overcome this problem. Carefully designed GPU code for image classification can be up to two orders of magnitude faster than its CPU counterpart  \cite{uetz:2009,strigl:2010}. Hence, to train huge DNN in hours or days, we implement them on GPU, building upon the work of \cite{Ciresan:2010,Ciresan:2011a}. The training algorithm is fully online, i.e. weight updates occur after each error back-propagation step.
We will show that properly trained big and deep DNNs can outperform all previous methods, and demonstrate that unsupervised  initialization/pretraining is not necessary (although we don't deny that it might help sometimes, especially for small datasets).  We also show how combining several DNN columns into a Multi-column DNN (MCDNN) further decreases the error rate by 30-40\%.

\section{Architecture}

The initially random weights of the DNN are iteratively trained to minimize the classification error on a set of labeled training images; generalization performance is then tested on a separate set of test images. Our architecture does this by combining several techniques in a novel way:

{\bf (1)}
Unlike the shallow NN used in many 1990s applications, ours are deep, inspired by the Neocognitron \cite{fukushima:1980}, with many (6-10) layers of non-linear neurons stacked on top of each other, comparable to the number of layers found between retina and visual cortex of macaque monkeys \cite{Bichot:05}.

{\bf (2)}
It was shown \cite{Hochreiter:01book} that such multi-layered DNN are hard to train by standard gradient descent \cite{Werbos:74,LeCun:85,Rumelhart:86}, the method of choice from a mathematical/algorithmic point of view. Today's computers, however, are fast enough for this, more than 60000 times faster than those of the early 90s\footnote{1991 486DX-33 MHz, 2011 i7-990X 3.46 GHz}. Carefully designed code for massively parallel graphics processing units (GPUs normally used for video games) allows for gaining an additional speedup factor of 50-100 over serial code for standard computers.
Given enough labeled data, our networks do not need additional heuristics such as unsupervised pre-training \cite{Salakhutdinov:2007,ranzato:2007,bengio:2007,erhan:2010} or carefully prewired synapses \cite{Riesenhuber:1999,Serre:2007}.

\begin{figure}[h!]
    \centering
    \setlength{\fboxsep}{10pt}
    \setlength{\fboxrule}{0pt}
    \subfigure[] {\fbox{\includegraphics[width=0.25\linewidth]{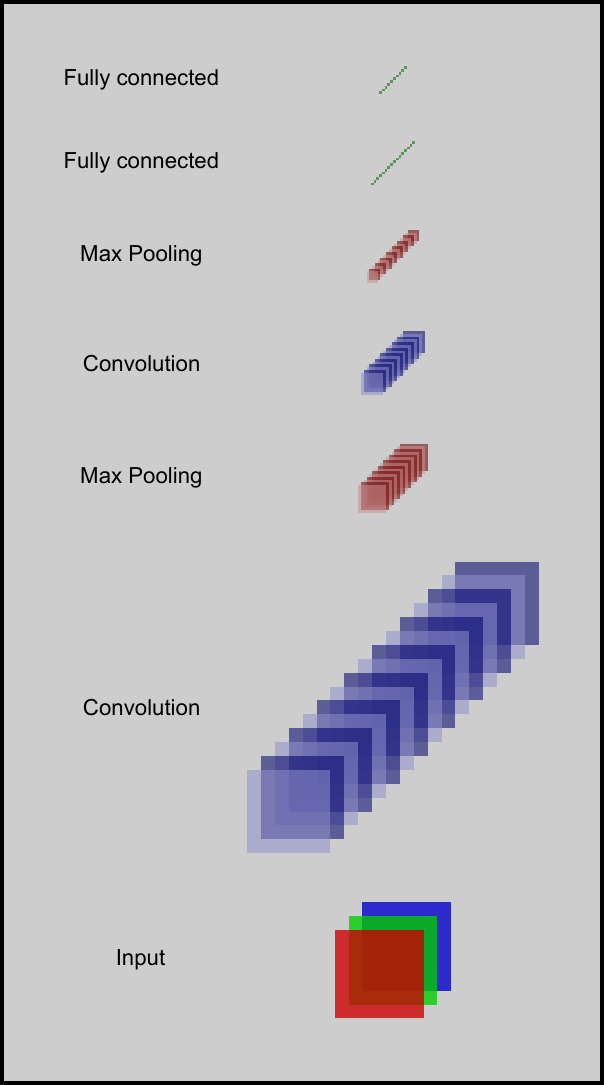}}}
    \subfigure[] {\fbox{\includegraphics[width=0.45\linewidth]{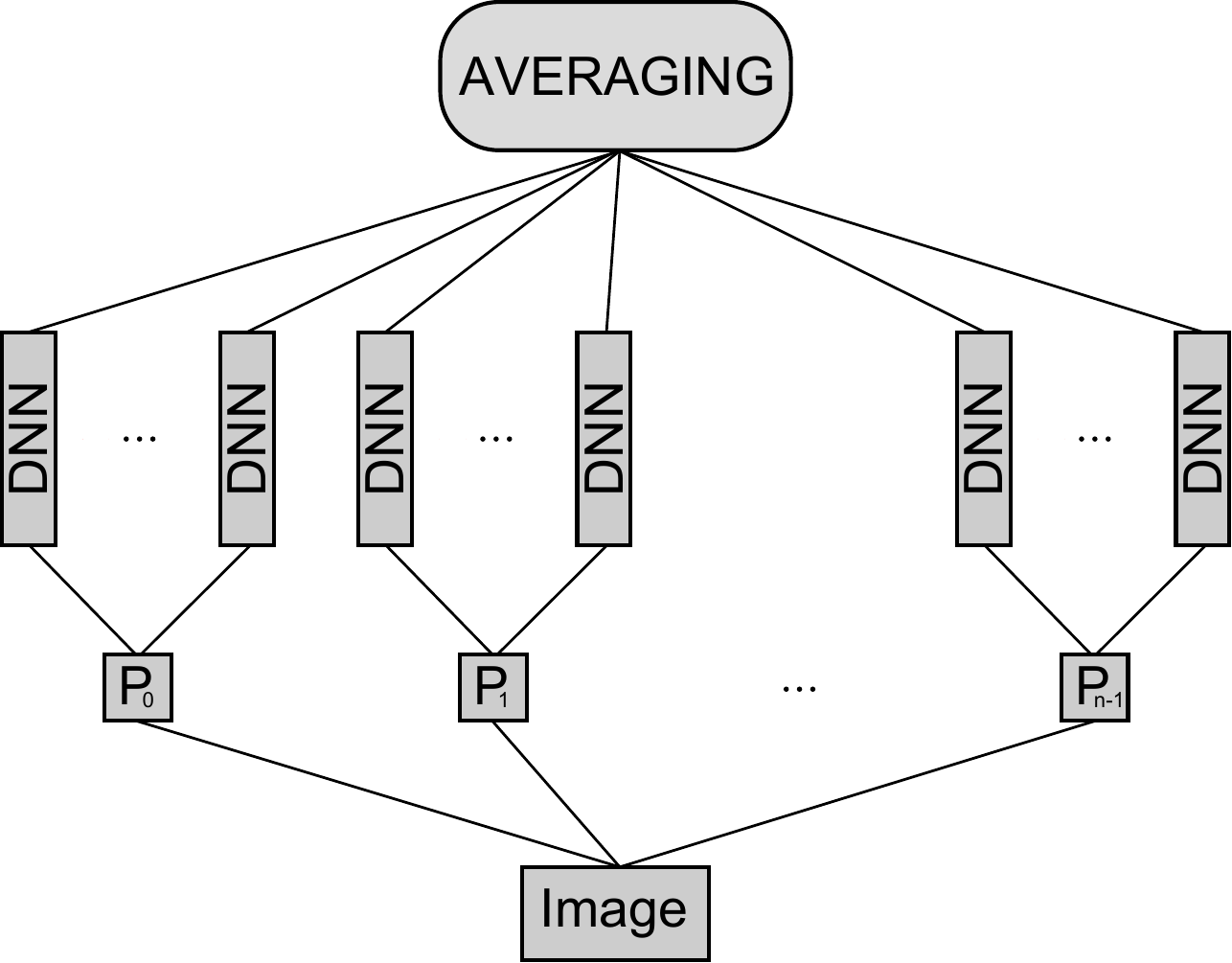}}}
\\
    \subfigure[] {\fbox{\includegraphics[width=0.6\linewidth]{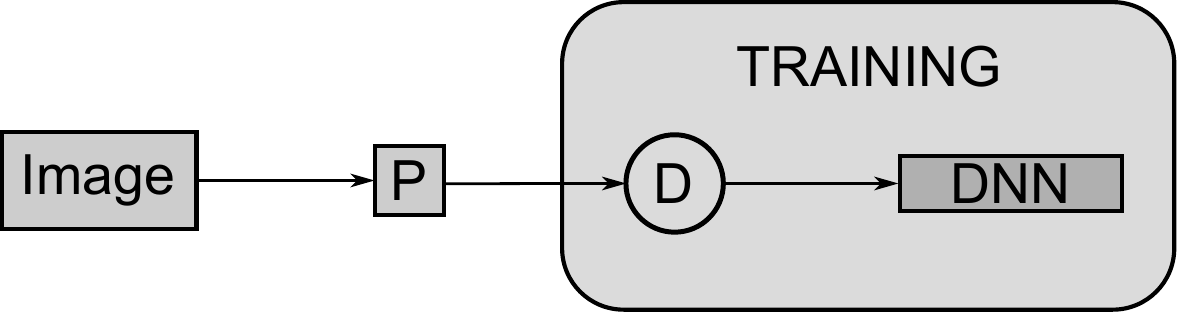}}}
    \caption{(a) DNN architecture. (b) MCDNN architecture. The input image can be preprocessed by $P_{0}-P_{n-1}$ blocks. An arbitrary number of columns can be trained on inputs preprocessed in different ways. The final predictions are obtained by averaging individual predictions of each DNN. (c) Training a DNN.
The dataset is preprocessed before training, then, at the beginning of every epoch, the images are distorted (D block). See text for more explanations.
}
\label{fig:MCDNN}
\end{figure}

{\bf (3)}
The DNN of this paper (Fig.~\ref{fig:MCDNN}a) have 2-dimensional layers of winner-take-all neurons \cite{Kohonen:88,WillshawMalsburg:76} with overlapping receptive fields whose weights are shared \cite{lecun:1998,behnke:2003,simard:2003,Ciresan:2011a}. Given some input pattern, a simple max pooling technique \cite{Riesenhuber:1999} determines winning neurons by partitioning layers into quadratic regions of local inhibition, selecting the most active neuron of each region. The winners of some layer represent a smaller, down-sampled layer with lower resolution, feeding the next layer in the hierarchy.
 The approach is inspired by Hubel and Wiesel's seminal work on the cat's primary visual cortex \cite{Wiesel:1959}, which identified orientation-selective \emph{simple cells} with overlapping local receptive fields and \emph{complex cells} performing down-sampling-like operations \cite{Hubel:1962}.

{\bf (4)}
Note that at some point down-sampling automatically leads to the first 1-dimensional layer.
From then on, only trivial 1-dimensional winner-take-all regions are possible, that is, the top part of the hierarchy
becomes a standard multi-layer perceptron (MLP) \cite{Werbos:74,LeCun:85,Rumelhart:86}.
Receptive fields and winner-take-all regions of our DNN often are (near-)minimal, e.g., only 2x2 or 3x3 neurons.
This results in (near-)maximal depth of  layers with non-trivial (2-dimensional) winner-take-all regions.
In fact, insisting on minimal 2x2 fields automatically defines the entire deep architecture,
apart from the number of different convolutional kernels per layer \cite{lecun:1998,behnke:2003,simard:2003,Ciresan:2011a}
and the depth of the plain MLP on top.

{\bf (5)}
Only winner neurons are trained, that is,
other neurons cannot forget what they learnt so far, although they may be affected by weight changes in more peripheral layers.
The resulting decrease of synaptic changes per time interval corresponds to biologically plausible reduction of energy consumption.
Our training algorithm is fully online, i.e. weight updates occur after each gradient computation step.

{\bf (6)}
Inspired by microcolumns of neurons in the cerebral cortex, we combine several DNN columns to form a Multi-column DNN (MCDNN). Given some input pattern, the predictions of all columns are democratically averaged. Before training, the weights (synapses) of all columns are randomly initialized. Various columns can be trained on the same inputs, or on inputs preprocessed in different ways.  The latter helps to reduce both error rate and number of columns required to reach a given accuracy. The MCDNN architecture and its training and testing procedures are illustrated in Figure \ref{fig:MCDNN}.

\section{Experiments}

In the following we give a detailed description of all the experiments we performed. We evaluate our architecture on various commonly used object recognition benchmarks and improve the state-of-the-art on all of them. The description of the DNN architecture used for the various experiments is given in the following way: 2x48x48-100C5-MP2-100C5-MP2-100C4-MP2-300N-100N-6N represents a net with 2 input images of size 48x48, a convolutional layer with 100 maps and 5x5 filters, a max-pooling layer over non overlapping regions of size 2x2, a convolutional layer with 100 maps and 4x4 filters, a max-pooling layer over non overlapping regions of size 2x2, a fully connected layer with 300 hidden units, a fully connected layer with 100 hidden units and a fully connected output layer with 6 neurons (one per class). We use a scaled hyperbolic tangent activation function for convolutional and fully connected layers, a linear activation function for max-pooling layers and a softmax activation function for the output layer. All DNN are trained using on-line gradient descent with an annealed learning rate. During training, images are continually translated, scaled and rotated (even elastically distorted in case of characters), whereas only the original images are used for validation. Training ends once the validation error is zero or when the learning rate reaches its predetermined minimum. Initial weights are drawn from a uniform random distribution in the range $[-0.05,0.05]$.

\subsection{MNIST}

\begin{figure}[h]
	\centering
	\setlength{\fboxsep}{10pt}
	\setlength{\fboxrule}{0pt}
	\subfigure[] {\fbox{\includegraphics[width=0.5\columnwidth]{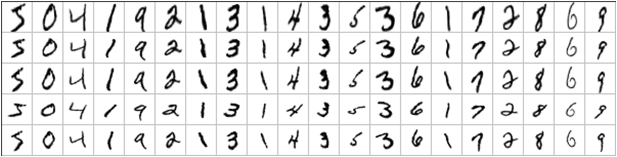}\label{fig:MNIST-distortions}}}
	\subfigure[] {\fbox{\includegraphics[width=0.4\columnwidth]{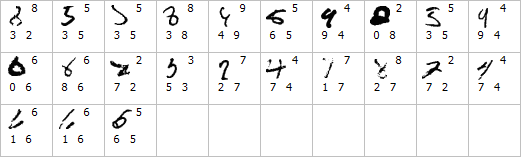}\label{fig:MNIST-errors}}}\\
	\subfigure[] {\fbox{\includegraphics[width=0.66\columnwidth]{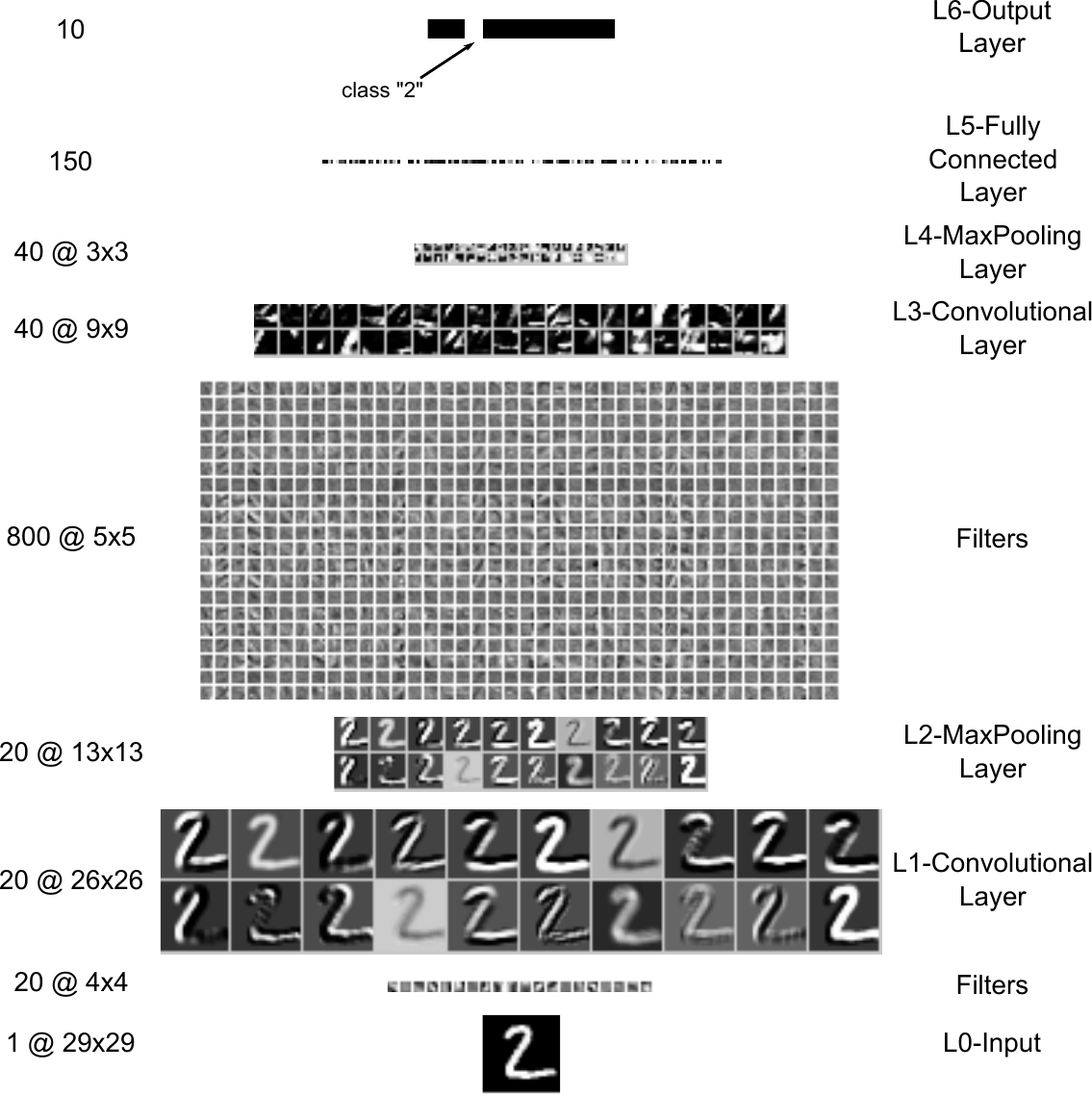}\label{fig:MNIST-DNN}}}\\
	
\caption{(a) Handwritten digits from the training set (top row) and their distorted versions after each epoch (second to fifth row). (b) The 23 errors of the MCDNN, with correct label (up right) and first and second best predictions (down left and right). (c) DNN architecture for MNIST. Output layer not drawn to scale; weights of fully connected layers not displayed.}
\label{fig:MNIST}
\end{figure}

The original MNIST digits  \cite{lecun:1998} are normalized such that the width or height of the bounding box equals 20 pixels. Aspect ratios for various digits vary strongly and we therefore create six additional datasets by normalizing digit width to 10, 12, 14, 16, 18, 20 pixels. This is like seeing the data from different angles. We train five DNN columns per normalization, resulting in a total of 35 columns for the entire MCDNN. All 1x29x29-20C4-MP2-40C5-MP3-150N-10N DNN are trained for around 800 epochs with an annealed learning rate (i.e. initialized with 0.001 multiplied by a factor of 0.993/epoch until it reaches 0.00003). Training a DNN takes almost 14 hours and after 500 training epochs little additional improvement is observed. During training the digits are randomly distorted before each epoch (see Fig. \ref{fig:MNIST}a for representative characters and their distorted versions \cite{Ciresan:2011a}). The internal state of a single DNN is depicted in Figure \ref{fig:MNIST}c, where a particular digit is forward propagated through a trained network and all activations together with the network weights are plotted.

Results of all individual nets and various MCDNN are summarized in Table \ref{tab:MNISTnets}. MCDNN of 5 nets trained with the same preprocessor achieve better results than their constituent DNNs, except for original images (Tab.~\ref{tab:MNISTnets}). 
The MCDNN has a very low 0.23\% error rate, improving state of the art by at least 34\%  \cite{Ciresan:2010,Ciresan:2011a, RanzatoCVPR:2007} (Tab.~\ref{tab:MNISTresults}). This is the first time an artificial method comes close to the $\approx$0.2\% error rate of humans on this task \cite{lecun-95a}. Many of the wrongly classified digits either contain broken or strange strokes, or have wrong labels. The 23 errors (Fig. \ref{fig:MNIST}b) are associated with 20 correct second guesses.

We also trained a single DNN on all 7 datasets simultaneously which yielded worse result (0.52\%) than both MCDNN and their individual DNN. This shows that the improvements come from the MCDNN and not from using more preprocessed data.

\begin{table*}[ht!]
	\caption{Test error rate [\%] of the 35 NNs trained on MNIST. Wxx - width of the character is normalized to xx pixels}
	\label{tab:MNISTnets}
	\small
	\centering
  \begin{tabular}{c|ccccccc}
 Trial		&	W10 		& 	W12		&	W14		&	W16		&	W18		&	W20		&	ORIGINAL	\\
   \hline
1 & 0.49 & 0.39 & 0.40 & 0.40 & 0.39 & 0.36 & 0.52\\
2 & 0.48 & 0.45 & 0.45 & 0.39 & 0.50 & 0.41 & 0.44\\
3 & 0.59 & 0.51 & 0.41 & 0.41 & 0.38 & 0.43 & 0.40\\
4 & 0.55 & 0.44 & 0.42 & 0.43 & 0.39 & 0.50 & 0.53\\
5 & 0.51 & 0.39 & 0.48 & 0.40 & 0.36 & 0.29 & 0.46\\
\hline\\
avg. & 0.52$\pm$0.05 & 0.44$\pm$0.05 & 0.43$\pm$0.03 & 0.40$\pm$0.02 & 0.40$\pm$0.06 & 0.39$\pm$0.08 & 0.47$\pm$0.05 \\
&  \multicolumn{7}{c}{35-net average error: 0.44$\pm$0.06}\\
\hline \\
5 columns	& 0.37 & 0.26 & 0.32 & 0.33 & 0.31 & 0.26 & 0.46\\
MCDNN\\
\hline\\
& \multicolumn{7}{c}{35-net MCDNN: {\bf 0.23\%}}\\
  \end{tabular}
\end{table*}

\begin{table}[h]
\caption{Results on MNIST dataset.} 
\label{tab:MNISTresults}
\small
\begin{center}
\begin{tabular}{c|cc}
Method		&	Paper				&	Error rate[\%]\\
\hline
CNN			&	\cite{simard:2003}		&	0.40		\\
CNN			&	\cite{ranzato:2006}		&	0.39		\\
MLP				&	\cite{Ciresan:2010}		&	0.35		\\
CNN committee	&	\cite{Ciresan:2011c}	&	0.27		\\
MCDNN		&	this					&	\bf{0.23}
\end{tabular}
\end{center}
\end{table}

How are the MCDNN errors affected by the number of preprocessors? We train 5 DNNs on all 7 datasets. A MCDNN '$y$ out-of-7' ($y$ from 1 to 7) averages $5y$ nets trained on $y$ datasets. Table~\ref{tab:MNISTpreprocessings} shows that more preprocessing results in lower MCDNN error.

We also train 5 DNN for each odd normalization, i.e. W11, W13, W15, W17 and W19. The 60-net MCDNN performs (0.24\%) similarly to the 35-net MCDNN, indicating that additional preprocessing does not further improve recognition.

\begin{table}[h!]
	\caption{Average test error rate [\%] of MCDNN trained on $y$ preprocessed datasets.}
	\small
	\centering
  \begin{tabular}{cc|c}
	$y$	& 	\# MCDNN		&	Average Error[\%]\\
	\hline
	1	&	7			&	0.33$\pm$0.07\\
	2	&	21			&	0.27$\pm$0.02\\
	3	&	35			&	0.27$\pm$0.02\\
	4	&	35			&	0.26$\pm$0.02\\
	5	&	21			&	0.25$\pm$0.01\\
	6	&	7			&	0.24$\pm$0.01\\
	7	&	1			&	0.23\\
  \end{tabular}
  	\label{tab:MNISTpreprocessings}
\end{table}

We conclude that MCDNN outperform DNN trained on the same data, and that different preprocessors further decrease the error rate.

\subsection{NIST SD 19}

The 35-columns MCDNN architecture and preprocessing used for MNIST are also applied to Latin characters from NIST SD 19 \cite{NIST}. For all tasks our MCDNN achieves recognition rates 1.5-5 times better than any published result (Tab.~\ref{tab:NISTSD19}). In total there are 82000 characters in the test set, but there are many more easy to classify digits (58000) than hard to classify letters (24000). This explains the lower overall error rate of the 62-class problem compared to the 52-class letters problem. From all errors of the 62-class problem 3\% of the 58000 digits are misclassified and 33\% of the 24000 letters are misclassified. Letters are in general more difficult to classify, but there is also a higher amount of confusion between similar lower- and upper-case letters such as i,I and o,O for example. Indeed, error rates for the case insensitive task drop from 21\% to 7.37\%. If the confused upper- and lower-case classes are merged, resulting in 37 different classes, the error rate is only slightly bigger (7.99\%). Upper-case letters are far easier to classify (1.83\% error rate) than lowercase letters (7.47\%) due to the smaller writer dependent in-class variability. For a detailed analysis of all the errors and confusions between different classes, the confusion matrix is most informative (Supplementary material Fig. S1).

\begin{table}[h]
\caption{Average error rates of MCDNN for all experiments, plus results from the literature. * case insensitive}
\label{tab:NISTSD19}
\small
\centering
\begin{tabular}{l|c|rr}
Data 			&  MCDNN			&\multicolumn{2}{c}{Published results}  \\
(task) 		&  error [\%]		&\multicolumn{2}{c}{Error[\%] and paper}\\
\hline
all (62)		&	{\bf11.63} 	& & \\
digits (10)		& 	{\bf0.77}		& 3.71 \cite{granger:2007}& 1.88 \cite{milgram:2005}  \\
letters (52)	& 	{\bf21.01}		& 30.91\cite{KoerichK:2005} & \\
letters* (26)	&  	{\bf7.37}		& 13.00 \cite{cavalin:2006} & 13.66\cite{KoerichK:2005} \\
merged (37)	&	{\bf7.99}		&  & \\ 
uppercase (26)	& 	{\bf1.83}		& 10.00 \cite{cavalin:2006} & 6.44 \cite{dossantos:2008} \\
lowercase (26)	& 	{\bf7.47}		& 16.00 \cite{cavalin:2006}& 13.27 \cite{KoerichK:2005} \\
\end{tabular}
\end{table} 

\subsection{Chinese characters}

Compared to Latin character recognition, isolated Chinese character recognition is a much harder problem, mainly because of the much larger category set, but also because of wide variability of writing styles, and the confusion between similar characters. We use a dataset from the Institute of Automation of Chinese
Academy of Sciences (CASIA \cite{cheng-lin:2010}), which contains 300 samples for each of 3755 characters (in GB1 set). This resulted in a data set with more than 1 Million characters (3GB of data) which posed a major computational challenge even to our system. Without our fast GPU implementation the nets on this task would train for more than one year. Only the forward propagation of the training set takes 27h on a normal CPU, and training a single epoch would consequently have lasted several days. On our fast GPU implementation on the other hand, training a single epoch takes 3.4h, which makes it feasible to train a net within a few days instead of many months. 

We train following DNN, 1x48x48-100C3-MP2-200C2-MP2-300C2-MP2-400C2-MP2-500N-3755N, on offline as well as on online characters. For the offline character recognition task, we resize all characters to 40x40 pixels and place them in the center of a 48x48 image. The contrast of each image is normalized independently. As suggested by the organizers, the first 240 writers from the database CASIA-HWDB1.1 are used for training and the remaining 60 writers are used for testing. The total numbers of training and test characters are 938679 and 234228, respectively.

For the online dataset, we draw each character from its list of coordinates, resize the resulting images to 40x40 pixels and place them in the center of a 48x48 image. Additionally, we smooth-out the resulting images with a Gaussian blur filter over a 3x3 pixel neighborhood and uniform standard deviation of 0.75. As suggested by the organizers, the characters of 240 writers from database CASIA-OLHWDB1.1 are used for training the classifier and the characters of the remaining 60 writers are used for testing. The resulting numbers of training and test characters are 939564 and 234800, respectively.

All methods previously applied to this dataset perform some feature extraction followed by a dimensionality reduction, whereas our method directly works on raw pixel intensities and learns the feature extraction and dimensionality reduction in a supervised way. On the offline task we obtain an error rate of {\bf 6.5\%} compared to 10.01\% of the best method \cite{cheng-lin:2010}. Even though much information is lost when drawing a character from it's coordinate sequence, we obtain a recognition rate of {\bf 5.61\%} on the online task compared to 7.61\% of the best method \cite{cheng-lin:2010}.

We conclude that on this very hard classification problem, with many classes (3755) and relatively few samples per class (240), our fully supervised DNN beats the current state-of-the-art methods by a large margin.

\subsection{Traffic signs}
Recognizing traffic signs is essential for the automotive industry's efforts in the field of driver's assistance, and for many other traffic-related applications. We use the GTSRB traffic sign dataset \cite{stallkamp:2011}. 

The original color images contain one traffic sign each, 
with a border of 10\% around the sign. They vary in size
from $15\times15$ to $250\times250$ pixels and are not necessarily
square. The actual traffic sign is not always centered within the
image; its bounding box is part of the
annotations. The training set consists of 26640 images; the test
set of 12569 images. We crop all images and 
process only within the bounding box. Our DNN implementation requires all
training images to be of equal size. After visual
inspection of the image size distribution  we
resize all images to $48\times48$ pixels. As a consequence, scaling
factors along both axes are different for traffic signs with
rectangular bounding boxes. Resizing forces them to have square
bounding boxes.

\begin{figure}[h]
	\centering
	\setlength{\fboxsep}{5pt}
	\setlength{\fboxrule}{0pt}
	\subfigure[] {\fbox{\includegraphics[width=0.3\linewidth]{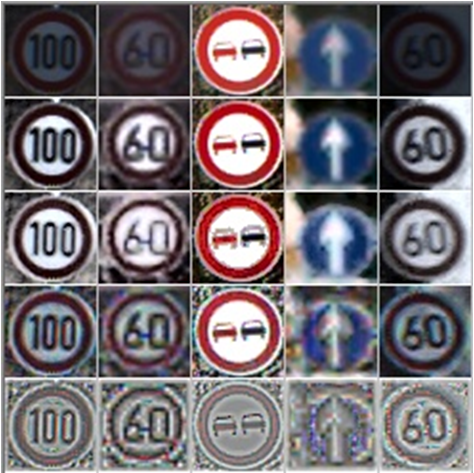}\label{fig:GTSRB-a}}}
	\subfigure[] {\fbox{\includegraphics[width=0.6\linewidth]{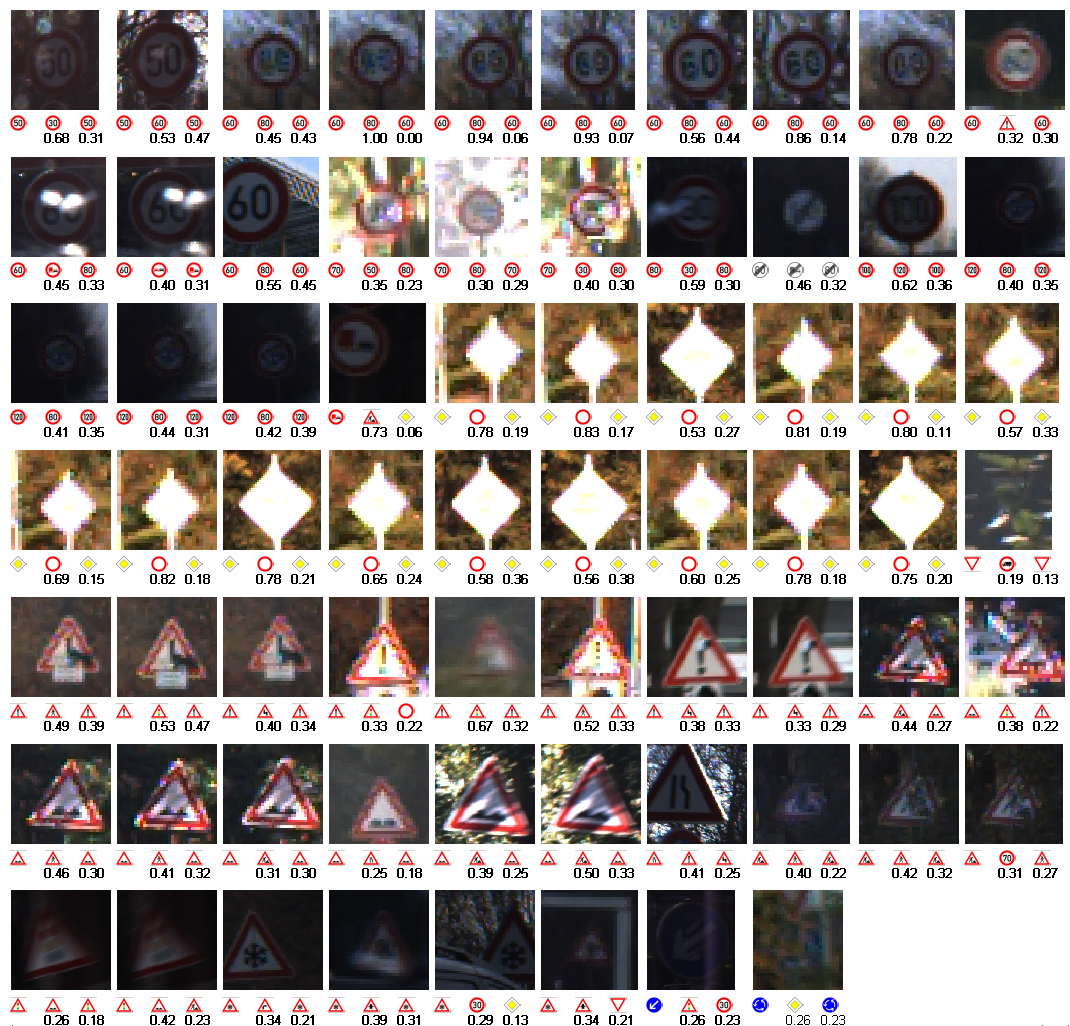}\label{fig:GTSRB-b}}}
	\caption{(a) Preprocessed images, from top to bottom: original, Imadjust, Histeq, Adapthisteq, Conorm. (b) The 68 errors of the MCDNN, with correct label (left) and first and second best predictions (middle and right).}

\label{fig:GTSRB}
\end{figure}

Our MCDNN is the only artificial method to outperform humans, who produced twice as many errors. Since traffic signs greatly vary in illumination and contrast, standard image preprocessing methods are used  to enhance/normalize them (Fig. \ref{fig:GTSRB}a and supplementary material). For each dataset five DNN are trained (architecture: 3x48x48-100C7-MP2-150C4-150MP2-250C4-250MP2-300N-43N), resulting in a MCDNN with 25 columns, achieving an error rate of {\bf 0.54\%} on the test set. Figure \ref{fig:GTSRB}b depicts all errors, plus ground truth and first and second predictions. Over 80\% of the 68 errors are associated with correct second predictions. Erroneously predicted class probabilities tend to be very low---here the MCDNN is quite unsure about its classifications. In general, however,  it is very confident---most of its predicted class probabilities are close to one or zero. Rejecting only 1\% percent of all images (confidence below 0.51) results in an even lower error rate of 0.24\%. To reach an error rate of 0.01\% (a single misclassification), only 6.67\% of the images have to be rejected (confidence below 0.94). Our method outperforms the second best algorithm by a factor of 3.  It takes 37 hours to train the MCDNN with 25 columns on four GPUs. The trained MCDNN can check 87 images per second on one GPU (and 2175 images/s/DNN).

\subsection{CIFAR 10}
CIFAR10 is a set of  natural color images of 32x32 pixels \cite{krizhevsky:2009}. It contains 10 classes, each with 5000 training samples and 1000 test samples. Images vary greatly within each class. They are not necessarily centered, may contain only parts of the object, and show different backgrounds. Subjects may vary in size by an order of magnitude (i.e., some images show only the head of a bird, others an entire bird from a distance). Colors and textures of objects/animals also vary greatly.
 
Our DNN input layers have three maps, one for each color channel (RGB).  We use a 10-layer architecture with very small kernels:  3x32x32-300C3-MP2-300C2-MP2-300C3-MP2-300C2-MP2-300N-100N-10N. Just like for MNIST, the initial learning rate 0.001 decays by a factor of 0.993 after every epoch. Transforming CIFAR color images to gray scale reduces input layer complexity but increases error rates. Hence we stick to the original color images. As for MNIST, augmenting the training set with randomly (by at most 5\%) translated  images greatly decreases the error from 28\% to 20\% (the NN-inherent local translation invariance by itself is not sufficient). By additional scaling (up to $\pm$15\%), rotation (up to $\pm5^\circ$), and up to $\pm$15\% translation, the individual net errors decrease by another 3\% (Tab.~\ref{tab:CIFAR10}). The above small maximal bounds prevent loss of too much information leaked beyond the $32\times32$ pixels rectangle. 

\begin{table}[!t]
\caption{Error rates, averages and standard deviations for 10 runs of a
10 layer DNN on the CIFAR10 test set. The nets in the first row are trained on preprocessed images (see  traffic sign preprocessing), whereas those in the second row are trained on original images.}
\label{tab:CIFAR10}
\footnotesize
\centering
\begin{tabular}{c|cccc|c}
preprocessing	& \multicolumn{4}{c}{errors for 8 runs [\%]}				&	mean[\%]			\\
\hline
yes 			&	16.47	&	19.20	&	19.72	&	20.31	&	$18.93\pm1.69$		\\
no 			&	15.63	&	15.85	&	16.13	&	16.05	&	$15.91\pm0.22$		\\
\hline
			& \multicolumn{4}{c}{8-net average error:	17.42$\pm$1.96\%}\\
\hline
			& \multicolumn{4}{c}{8-net MCDNN error:	\bf{11.21\%}}\\
\hline
			&	 \multicolumn{5}{c}{previous state of the art: 18.50\% - \cite{CoatesICML:2011}; 19.51\% - \cite{Ciresan:2011a}}\\
\end{tabular}
\end{table}

\begin{figure}[ht!]
\hfill
\begin{center}
\includegraphics[width=0.7\linewidth]{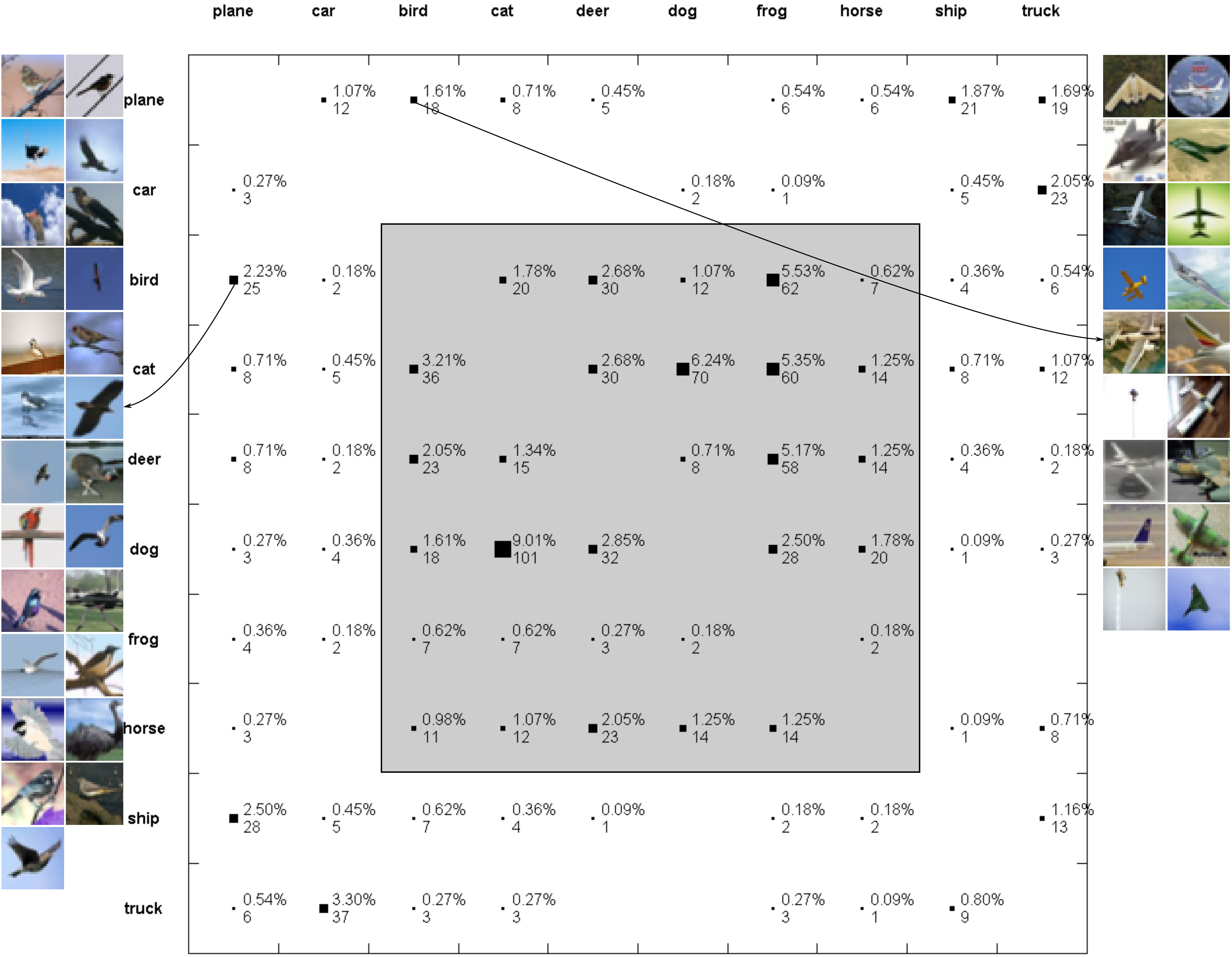}
\end{center}
\caption{Confusion matrix for the CIFAR10 MCDNN: correct labels on vertical axis; detected labels on horizontal axis. Square areas are proportional to error numbers, shown both as relative percentages of the total error number, and in absolute value. Left - images of all birds classified as planes. Right - images of all planes classified as birds. Confusion sub-matrix for animal classes has a gray backround.}
\label{Fig:CIFAR10confmat}
\end{figure}

We repeat the experiment with different random initializations and compute mean and standard deviation of the error, which is rather small for original images, showing that our DNN are robust. Our MCDNN obtains a very low error rate of 11.21\%, greatly rising the bar for this benchmark.

The confusion matrix (Figure~\ref{Fig:CIFAR10confmat}) shows that the MCDNN almost perfectly separates animals from artifacts, except for planes and birds, which seems natural, although humans easily distinguish almost all the incorrectly classified images, even if many are cluttered or contain only parts of the objects/animals  (see false positive and false negative images in Figure~\ref{Fig:CIFAR10confmat}). There are many confusions between different animals; the frog class collects most false positives from other animal classes, with very few false negatives. As expected, cats are hard to tell from dogs, collectively causing 15.25\% of the errors.

The MCDNN with 8 columns (four trained on original data and one trained for each preprocessing used also for traffic signs) reaches a low 11.21\% error rate, far better than any other algorithm.

\subsection{NORB}
\begin{figure}[ht!]
\hfill
\begin{center}
\includegraphics[width=0.5\linewidth]{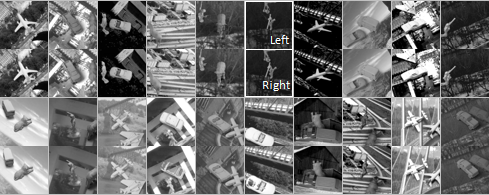}
\end{center}
\caption{Twenty NORB stereo images (left image - up, right image - down).}
\label{Fig:NORBsamples}
\end{figure}

We test a MCDNN with four columns on NORB (jittered-cluttered) \cite{lecun:2004}, a collection of stereo images of 3D models (Figure~\ref{Fig:NORBsamples}). The objects are centrally placed on randomly chosen backgrounds, and there is also cluttering from a peripherally placed second object. 
This database is designed for experimenting with 3D object recognition from shape. It contains images of 50 toys belonging to 5 generic categories: four-legged animals, human figures, airplanes, trucks, and cars. The objects were imaged by two cameras under 6 lighting conditions, 9 elevations (30 to 70 degrees every 5 degrees), and 18 azimuths (0 to 340 every 20 degrees). The training set has 10 folds of 29160 images each for a total of 291600 images; the testing set consists of two folds totalizing 58320 images.

No preprocessing is used for this dataset.  We scale down images from the original 108x108 to 48x48 pixels. This size is big enough to preserve the details present in images and small enough to allow fast training. We perform two rounds of experiments, using only the first two folds (to compare with previous results that do not use the entire training data) and using all training data.

We tested several distortion parameters with small nets and found that maximum rotation of $15^\circ$, maximum translation of 15\% and maximum scaling of 15\% are good choices, hence we use them for all NORB experiments.

To compare to previous results, we first train only on the first 2-folds of the data. The net architecture is deep, but has few maps per layer:  2x48x48-50C5-MP2-50C5-MP2-50C4-MP2-300N-100N-6N. The learning rate setup is: eta start 0.001; eta factor 0.95; eta stop 0.000003. Due to small net size, training is fast at 156s/epoch for 114 epochs. Testing one sample requires 0.5ms.  Even when we use less data to train, the MCDNN greatly improves the state of the art from 5\% to 3.57\% (Table~\ref{tab:NORBresults}).

Our method is fast enough to process the entire training set though. We use the same architecture but double the number of maps when training with all 10 folds: 2x48x48-100C5-MP2-100C5-MP2-100C4-MP2-300N-100N-6N. The learning rate setup remains the same. Training time increases to 34min/epoch because the net is bigger, and we use five times more data. Testing one sample takes 1.3ms. All of this pays off, resulting in a very low 2.70\% error rate, further improving the state of the art.

\begin{table}[!t]
\caption{Error rates, averages and standard deviations over 4 runs of a 9 layer DNN on the NORB  test set.}
\label{tab:NORBresults}
%\footnotesize
\centering
\begin{tabular}{c|cccc|c}
training	& \multicolumn{4}{c}{errors for 4 runs [\%]}				&	mean[\%]			\\
set size	&  \multicolumn{4}{c}{}		&				\\
\hline
first			&	4.49		&	4.71		&	4.82		&	4.85		&	$4.72\pm0.16$		\\
2 folds		& 	\multicolumn{5}{c}{4-net MCDNN error:	3.57\%}\\
\hline
all 			&	3.32		&	3.18		&	3.73		&	3.36		&	$3.40\pm0.23$		\\
10 folds		& 	\multicolumn{5}{c}{4-net MCDNN error:	{\bf 2.70\%}}\\
\hline
\hline
 \multicolumn{6}{c}{previous state of the art: 5.00\% - \cite{CoatesICML:2011}; 5.60\% - \cite{scherer:2010}}\\
\end{tabular}
\end{table}

Although NORB has only six classes, training and test instances sometimes differ greatly, making classification hard. More than 50\% of the errors are due to confusions between cars and trucks.
Considering second predictions, too, the error rate drops from 2.70\% to 0.42\%, showing that 84\% of the errors are associated with
a correct second prediction.

\section{Conclusion}
This is the first time human-competitive results are reported on widely used computer vision benchmarks. On many other image classification datasets our MCDNN improves the state-of-the-art by 30-80\%  (Tab.~\ref{tab:improvements}). We drastically improve recognition rates on MNIST, NIST SD 19, Chinese characters, traffic signs, CIFAR10 and NORB. Our method is fully supervised and does not use any additional unlabeled data source. Single DNN already are sufficient to obtain new state-of-the-art results; combining them into MCDNNs yields further dramatic performance boosts.

\begin{table}[h]
\caption{Results and relative improvements on different datasets.} 
%\footnotesize
\begin{center}
\begin{tabular}{c|ccc}
Dataset	&	Best result		&	MCDNN					&	Relative \\
		&	 of others [\%]	&	[\%]						&	improv. [\%]\\
\hline
MNIST	&	0.39			&	0.23						&	41		\\
NIST SD 19&	see Table~\ref{tab:NISTSD19}			&	see Table~\ref{tab:NISTSD19}			&	30-80	\\
HWDB1.0 on.	&	7.61		&	5.61						&	26		\\
HWDB1.0 off.	&	10.01		&	6.5						&	35		\\
%(chinese characters)&&&\\
CIFAR10	&	18.50		&	11.21					&	39		\\
traffic signs&	1.69			&	0.54						&	72		\\
NORB	&	5.00			&	2.70						&	46		\\
\end{tabular}
\label{tab:improvements}
\end{center}
\end{table}

\section*{Acknowledgment}

This work was partially supported by a FP7-ICT-2009-6 EU Grant under Project Code 270247: A Neuro-dynamic Framework for Cognitive Robotics: Scene Representations, Behavioral Sequences, and Learning.

\bibliographystyle{ieee}
\bibliography{bib}

%%%%%%%%%%%%%%%%%%%%%%%%%%%%%%%%%%%%%%%%%%%%%%%%%%%%%%%%%%%%%%%%%%%%%
%%%%%%%%%%%%%%%%%%%%%%%%%%%%%%%%%%%%%%%%%%%%%%%%%%%%%%%%%%%%%%%%%%%%%
%%%%%%%%%%%%%%%%%%%%%%%%%%%%%%%%%%%%%%%%%%%%%%%%%%%%%%%%%%%%%%%%%%%%%
%%%%%%%%%%%%%%%%%%%%%%%%%%%%%%%%%%%%%%%%%%%%%%%%%%%%%%%%%%%%%%%%%%%%%
\newpage
\section{Supplementary Material}

\subsection{Experiment details}

\subsubsection{NIST SD 19}

\begin{figure}[ht!]
\hfill
\begin{center}
\includegraphics[width=0.8\linewidth]{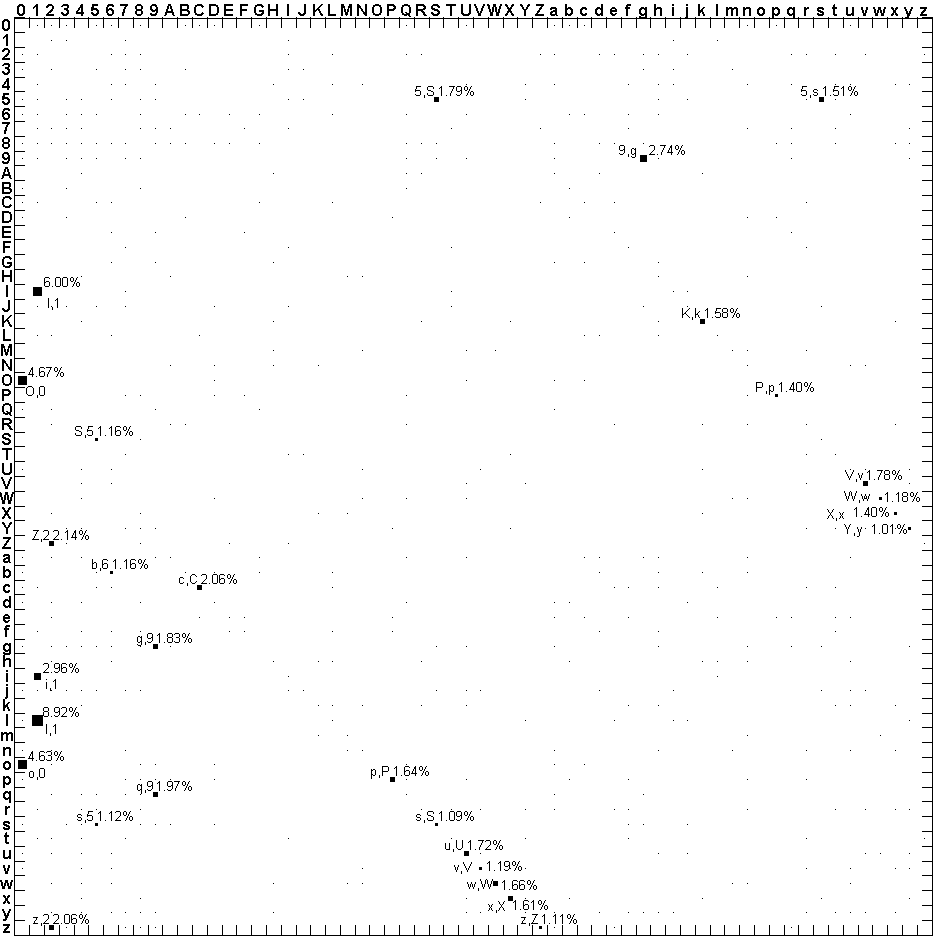}
\end{center}
\caption{Confusion matrix of the NIST SD 19 MCDNN trained on the 62-class task: correct labels on vertical axis; detected labels on horizontal axis. Square areas are proportional to error numbers, shown as relative percentages of the total error number. For convenience, class labels are written beneath the errors. Errors below 1\% of the total error number are not detailed.}
\label{Fig:confMatNISTAllCharacters}
\end{figure}

The confusion matrix of the 62 characters task (Fig.~\ref{Fig:confMatNISTAllCharacters}) shows that most of the errors are due to confusions between digits and letters and between lower- and upper-case letters.

\begin{figure}[ht!]
\hfill
\begin{center}
\includegraphics[width=0.3\linewidth]{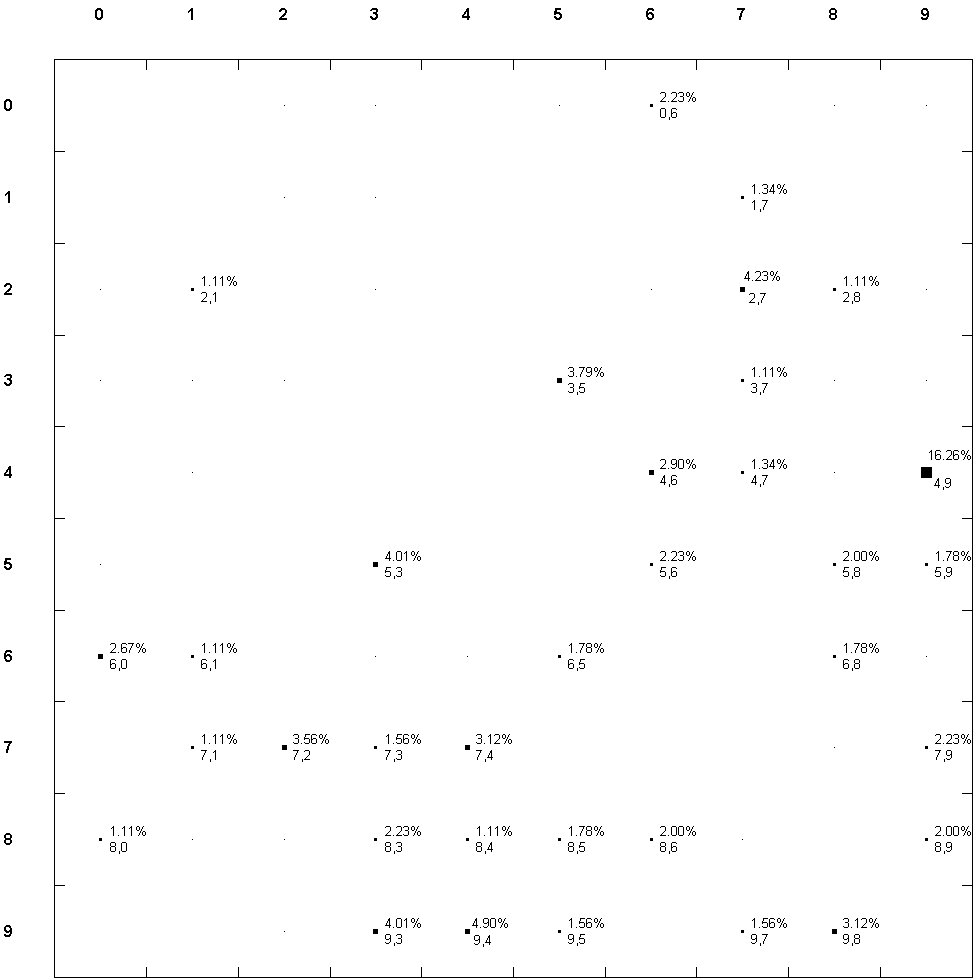}
\end{center}
\caption{Confusion matrix for the NIST SD 19 MCDNN trained on the digit task: correct labels on vertical axis; detected labels on horizontal axis. Square areas are proportional to error numbers, shown as relative percentages of the total error number. Class labels are shown beneath the errors. Errors below 1\% of the total error number are shown as dots without any details.}
\label{Fig:confMatNISTDigits}
\end{figure}

Not very surprisingly, the confusion matrix for the digit task (Fig. ~\ref{Fig:confMatNISTDigits}) shows that confusions between fours and nines are the most common error source.

\begin{figure}[ht!]
\hfill
\begin{center}
\includegraphics[width=0.65\linewidth]{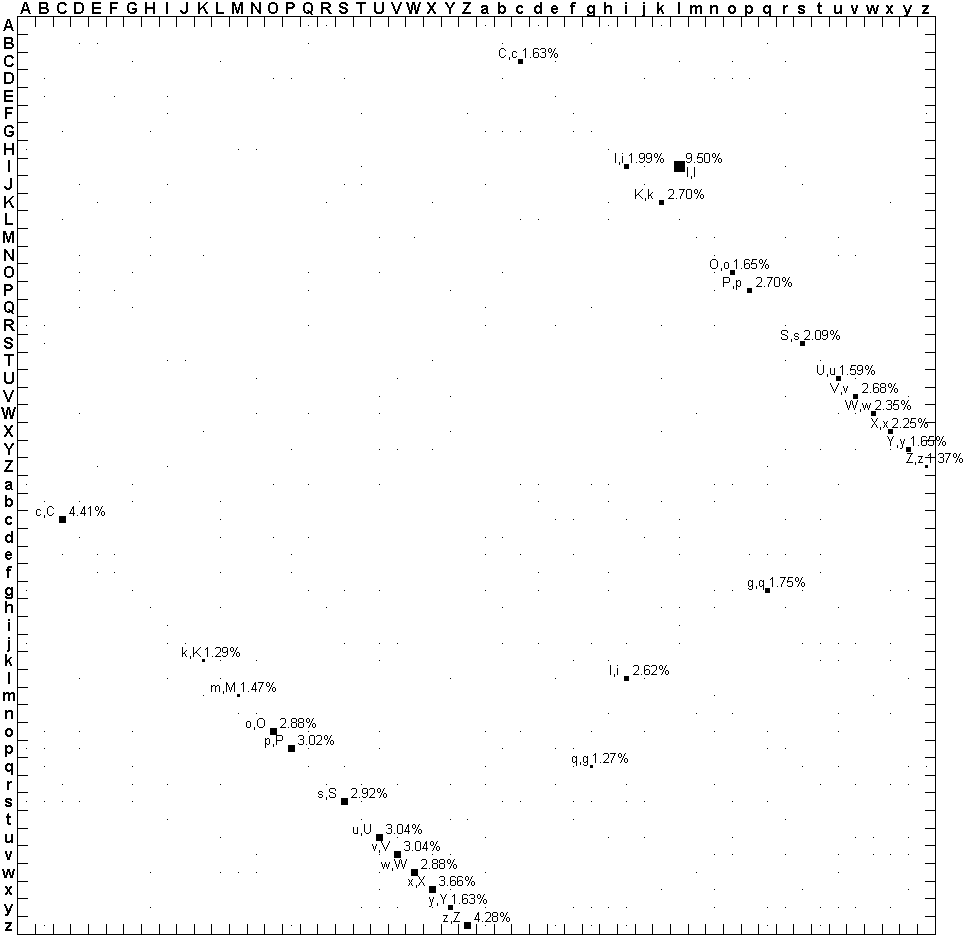}
\end{center}
\caption{Confusion matrix for the NIST SD 19 MCDNN trained on all 52 letters: correct labels on vertical axis; detected labels on horizontal axis. Square areas are proportional to error numbers, shown as relative percentages of the total error number. Class labels are shown beneath the errors. Errors below 1\% of the total error number are shown as dots without any details.}
\label{Fig:confMatNISTAllLetters}
\end{figure}

For the 52 letter task (case sensitive) the confusion matrix (Fig.~\ref{Fig:confMatNISTAllLetters}) shows that the MCDNN has mainly problems with upper- and lower-case confusions of the same letter. Other hard-to-distinguish classes are: 'q' and 'g', 'l' and 'i'.

\begin{figure}[ht!]
\hfill
\begin{center}
\includegraphics[width=0.5\linewidth]{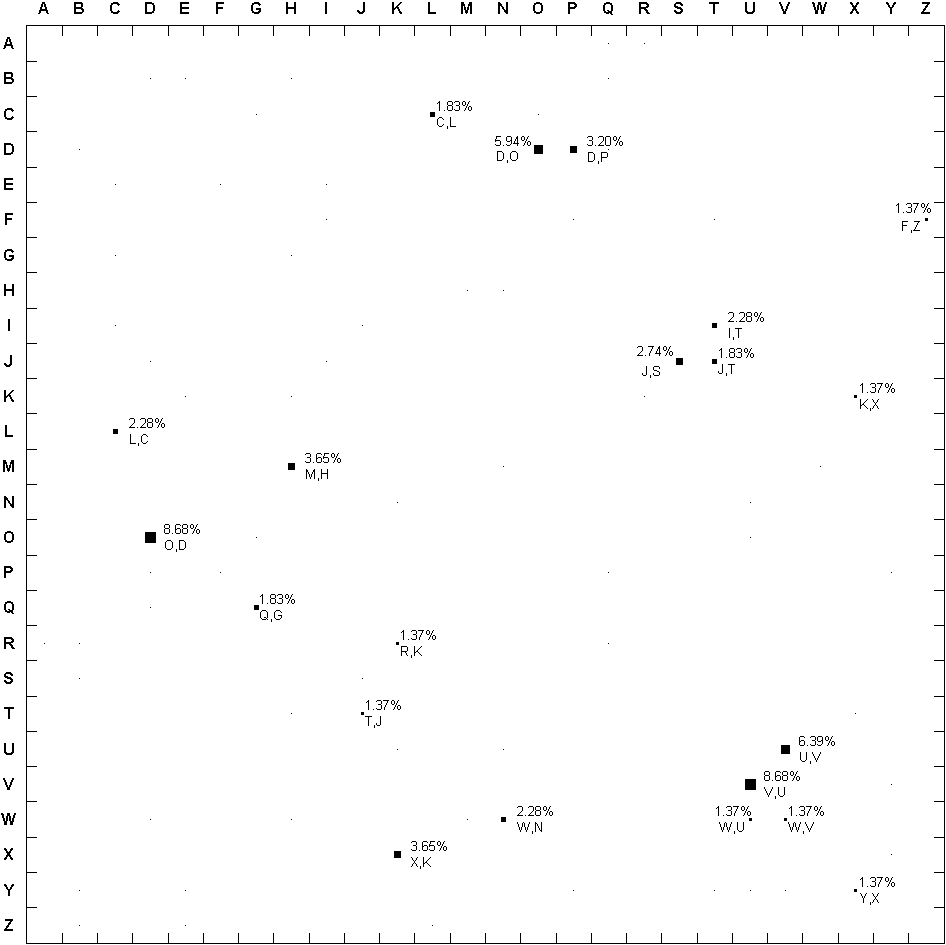}
\end{center}
\caption{Confusion matrix for the NIST SD 19 MCDNN trained on uppercase letters: correct labels on vertical axis; detected labels on horizontal axis. Square areas are proportional to error numbers, shown as relative percentages of the total error number. Class labels are shown beneath the errors. Errors below 1\% of the total error number are shown as dots without any details.}
\label{Fig:confMatNISTBigLetters}
\end{figure}

For the upper-case letter task the confusion matrix (Figure~\ref{Fig:confMatNISTBigLetters}) shows that the MCDNN has problems with letters of similar shape, i.e. 'D', and 'O', 'V' and 'U' etc. The total error of 1.82\% is very low though.

\begin{figure}[ht!]
\hfill
\begin{center}
\includegraphics[width=0.5\linewidth]{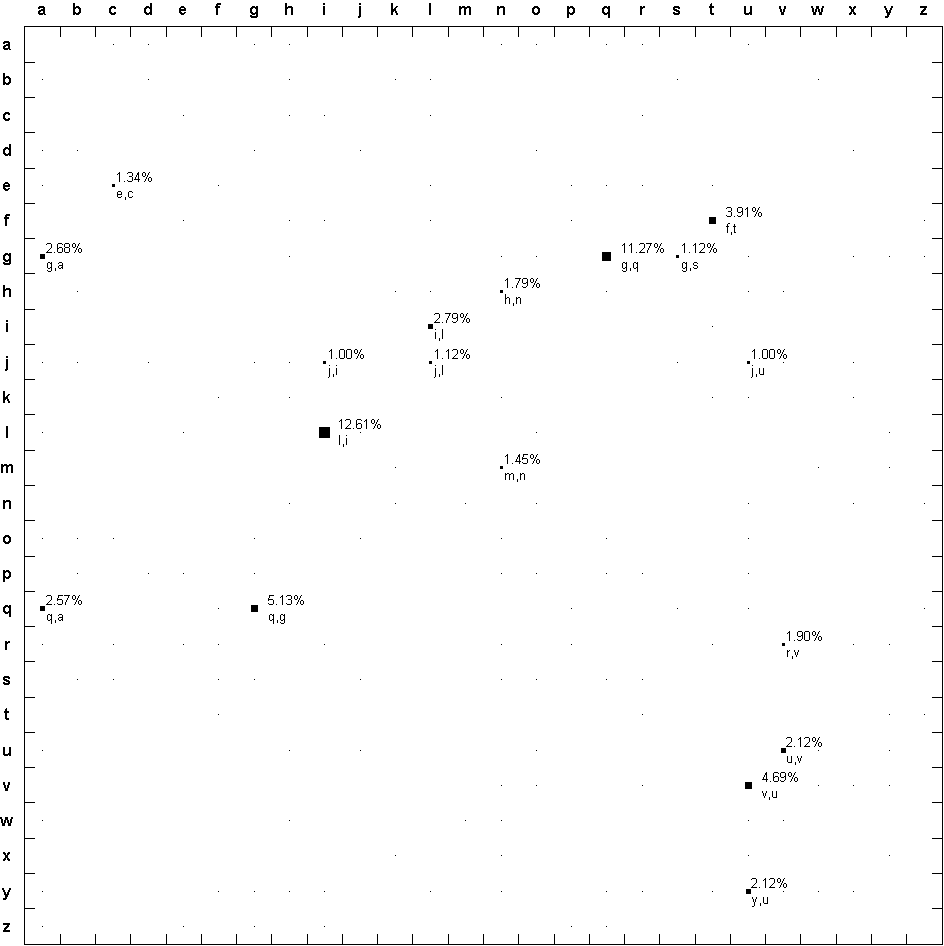}
\end{center}
\caption{Confusion matrix for the NIST SD 19 MCDNN trained on lowercase letters: correct labels on vertical axis; detected labels on horizontal axis. Square areas are proportional to error numbers, shown as relative percentages of the total error number. Class labels are shown beneath the errors. Errors below 1\% of the total error number are shown as dots without any details.}
\label{Fig:confMatSmallLetters}
\end{figure}

For the lower-case letter task the confusion matrix (Fig.~\ref{Fig:confMatSmallLetters}) shows that like with upper-case letters, the MCDNN has problems with letters of similar shapes, i.e. 'g', and 'q', 'v' and 'u' etc. But the total error is much higher (7.47\%) than for the upper-case letters task.

\begin{figure}[ht!]
\hfill
\begin{center}
\includegraphics[width=0.5\linewidth]{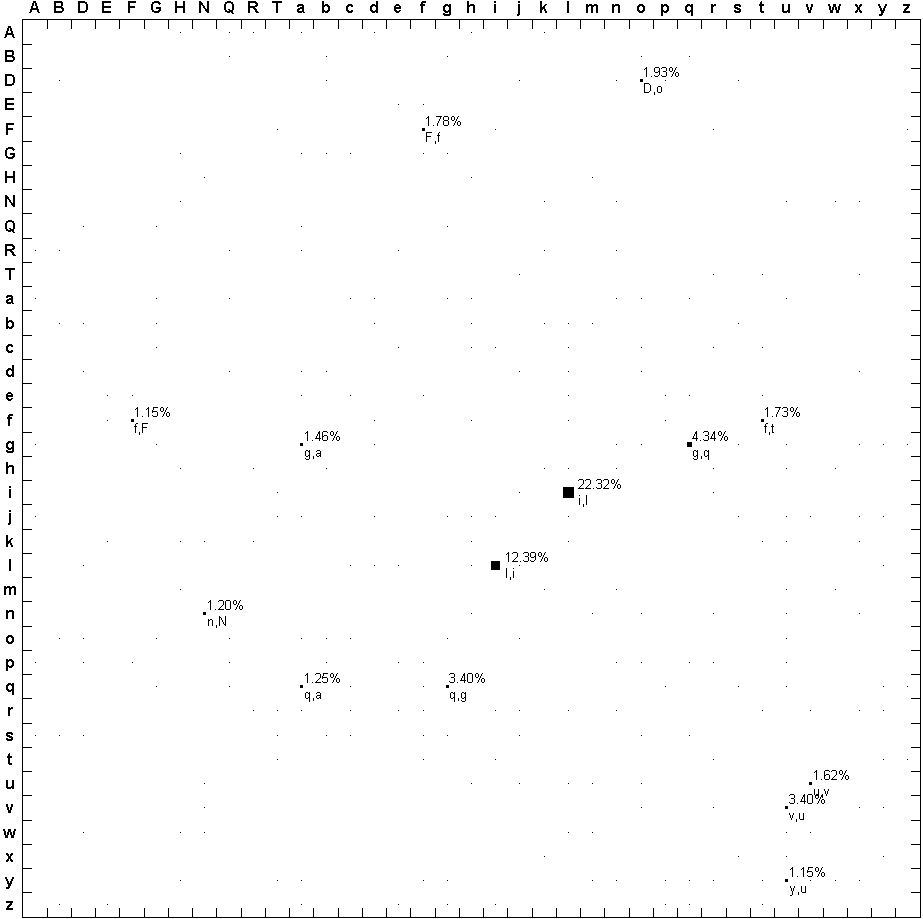}
\end{center}
\caption{Confusion matrix for the NIST SD 19 MCDNN trained on merged letters (37 classes): correct labels on vertical axis; detected labels on horizontal axis. Square areas are proportional to error numbers, shown as relative percentages of the total error number. Class labels are shown beneath the errors. Errors below 1\% of the total error number are shown as dots without any details.}
\label{Fig:confMatMergedLetters}
\end{figure}

For the merged-case letter task (37 classes) the confusion matrix (Figure~\ref{Fig:confMatMergedLetters}) shows that the MCDNN has mostly problems with letters of similar shapes, i.e. 'l', and 'i'. All upper-lower-case confusions of identical letters from the 52 class task vanish, the error shrinks by a factor of almost three down to 7.99\%.

The experiments on different subsets of the 62 character task clearly show that it is very hard to distinguish between small and capital letters. Also, digits 0 and 1 are hard to separate from letters O and I. Many of these problems could be alleviated by incorporating context  where possible.

%%%%%%%%%%%%%%%%%%%%%%%%%%%%%%%%%%%%%%%%%%%%%%%%%%%%%%%%%%%%%%%%%
\subsubsection{Traffic signs}

High contrast variation among the images calls for normalization. We test the following standard contrast normalizations:

\begin{itemize}
\item
{\bf Image Adjustment (Imadjust)} increases image contrast by mapping pixel intensities to new values such that 1\% of the data is saturated at low and high intensities \cite{matlab:2010}.
\item
{\bf Histogram Equalization (Histeq)} enhances contrast by transforming pixel intensities such that the output image histogram is roughly uniform \cite{matlab:2010}.
\item
{\bf Adaptive Histogram Equalization (Adapthisteq)} operates (unlike Histeq) on tiles rather than the entire image, we tiled the image in 8 nonoverlapping regions of 6x6 pixels. Each tile's contrast is enhanced such that its histogram becomes roughly uniform \cite{matlab:2010}.
\item
{\bf Contrast Normalization (Conorm)} enhances edges, filtering the input image by a difference of Gaussians, using a filter size of 5x5 pixels \cite{Sermanet:2011}. 
\end{itemize}

\begin{figure*}[ht!]
\hfill
\begin{center}
\includegraphics[width=\linewidth]{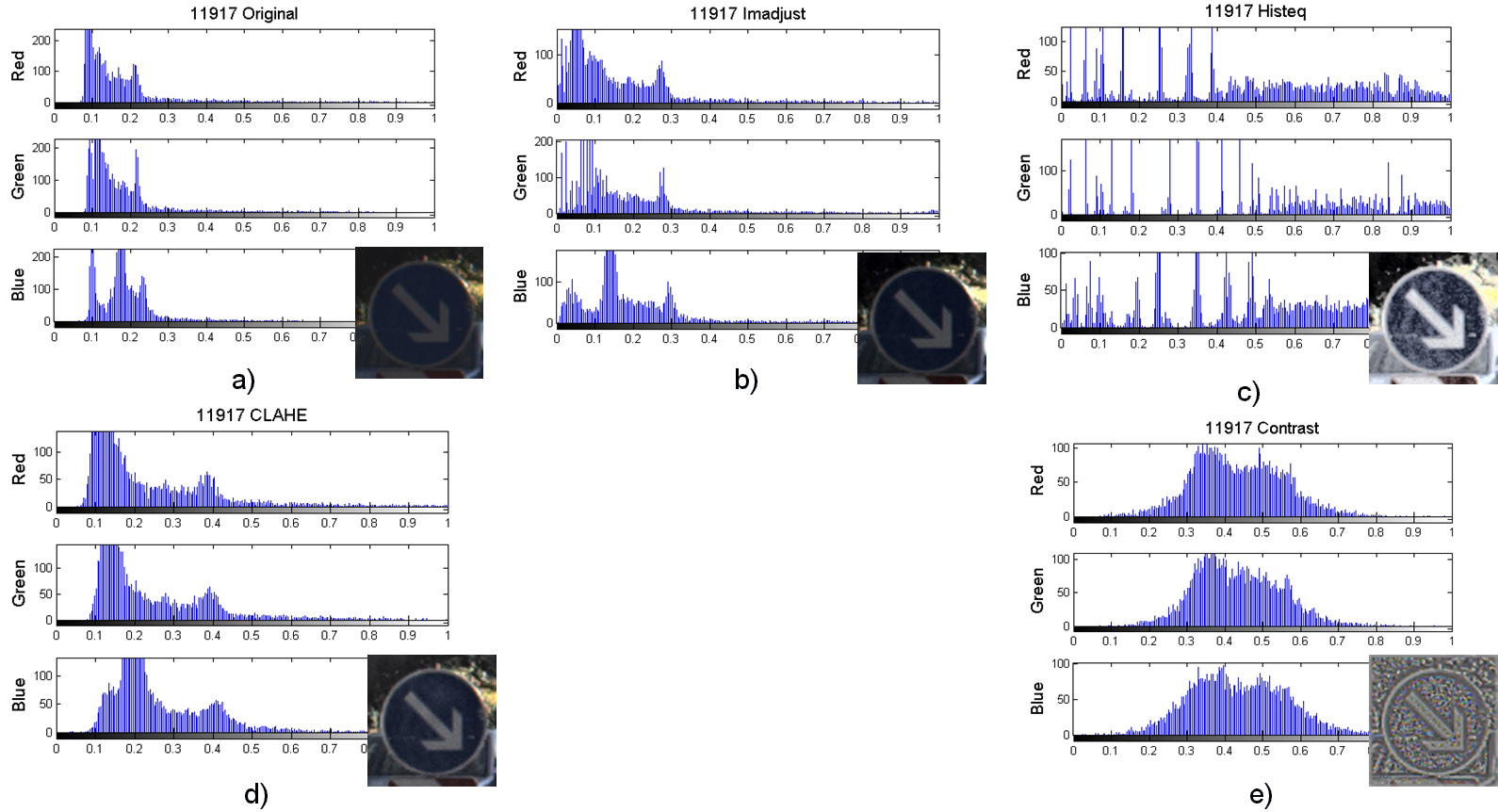}
\end{center}
\caption{Preprocessing.}
\label{Fig:GTSRBpreprocessing}
\end{figure*}

Note that the above normalizations, except Conorm, are performed in a color space with image intensity as one of its components. For this purpose we transform the image from RGB- to Lab-space, then perform normalization, then transform the result back to RGB-space. The effect of the four different normalizations is summarized in Figure \ref{Fig:GTSRBpreprocessing}, where histograms of pixel intensities together with original and normalized images are shown.

The DNN have three maps for the input layer, one for each color channel (RGB). The rest of the net architecture is detailed in Table~\ref{tab:netGTSRB}. We use a 10-layer architecture with very small max-pooling kernels.

\begin{table}[h!]
\caption{10 layer DNN architecture used for recognizing traffic signs.} 
\small
\begin{center}
\begin{tabular}{c|c|c|c}
Layer		&	Type			&	\# maps \& neurons			&	kernel	\\
\hline
0		&	input			&	3 maps of 48x48 neurons		&			\\
1		&	convolutional	&	100 maps of 42x42 neurons		&	7x7		\\
2		&	max pooling	&	100 maps of 21x21 neurons		&	2x2		\\
3		&	convolutional	&	150 maps of 18x18 neurons		&	4x4		\\
4		&	max pooling	&	150 maps of 9x9 neurons		&	2x2		\\
5		&	convolutional	&	250 maps of 6x6 neurons		&	4x4		\\
6		&	max pooling	&	250 maps of 3x3 neurons		&	2x2		\\
9		&	fully connected	&	300 neurons				&	1x1		\\
10		&	fully connected	&	43 neurons				&	1x1		\\
\end{tabular}
\label{tab:netGTSRB}
\end{center}
\end{table}

%%%%%%%%%%%%%%%%%%%%%%%%%%%%%%%%%%%%%%%%%%%%%%%%%%%%%%%%%%%%%%%%%
\newpage
\subsubsection{CIFAR10}
\begin{table}[h!]
\caption{10 layer DNN architecture used for CIFAR 10.} 
\small
\begin{center}
\begin{tabular}{c|c|c|c}
Layer		&	Type			&	\# maps \& neurons			&	kernel	\\
\hline
0		&	input			&	3 maps of 32x32 neurons		&			\\
1		&	convolutional	&	300 maps of 30x30 neurons		&	3x3		\\
2		&	max pooling	&	300 maps of 15x15 neurons		&	2x2		\\
3		&	convolutional	&	300 maps of 14x14 neurons		&	2x2		\\
4		&	max pooling	&	300 maps of 7x7 neurons		&	2x2		\\
5		&	convolutional	&	300 maps of 6x6 neurons		&	2x2		\\
6		&	max pooling	&	300 maps of 3x3 neurons		&	2x2		\\
7		&	convolutional	&	300 maps of 2x2 neurons		&	2x2		\\
8		&	max pooling	&	300 maps of 1x1 neurons		&	2x2		\\
9		&	fully connected	&	300 neurons				&	1x1		\\
10		&	fully connected	&	100 neurons				&	1x1		\\
11		&	fully connected	&	10 neurons				&	1x1		\\
\end{tabular}
\label{tab:netCIFAR10}
\end{center}
\end{table}

%%%%%%%%%%%%%%%%%%%%%%%%%%%%%%%%%%%%%%%%%%%%%%%%%%%%%%%%%%%%%%%%%%%
\subsubsection{NORB}

\begin{figure}[ht!]
\hfill
\begin{center}
\includegraphics[width=0.5\linewidth]{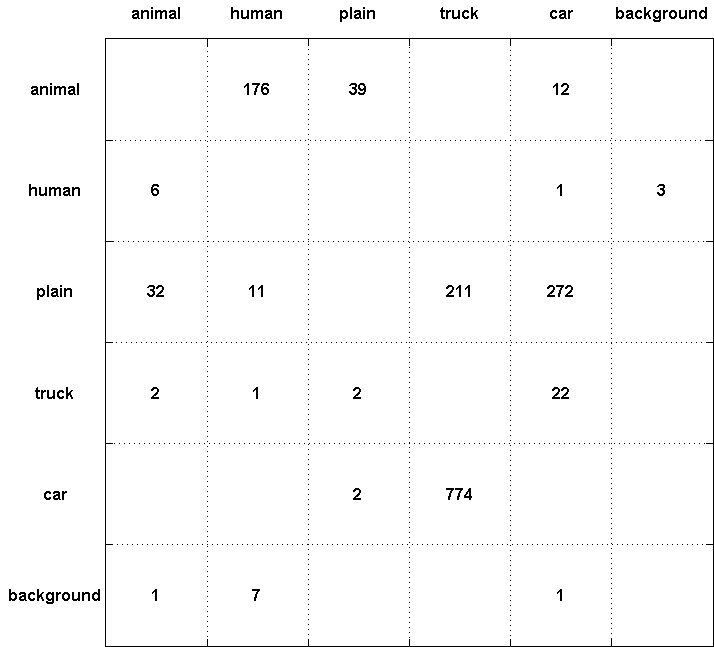}
\end{center}
\caption{Confusion matrix for the NORB: correct labels on vertical axis; detected labels on horizontal axis.}
\label{Fig:confMatNORB}
\end{figure}

\end{document}